\documentclass[letterpaper]{article}
\usepackage{aaai2027}
\usepackage[hyphens]{url}
\usepackage{graphicx}
\usepackage{natbib}
\usepackage{caption}
\usepackage{booktabs}
\usepackage{amsmath}

\nocopyright

\title{SMO: Compiling Natural Language Skills into Graphs of Reusable\\State-Move-Outcome Bricks}
\author{Chao Chen, Jiashuo Zhang, Yunhang Shen, Liangyu Lian, Rongrong Ji\corresponding}
\affiliations{Xiamen University}

\begin{document}

\maketitle

\begin{abstract}
Large language model agents increasingly rely on reusable natural-language skills, yet growing skill collections remain accumulations of isolated text artifacts that are difficult to compare, deduplicate, and compose.
We argue that the reusable content of a skill lies not in its surface wording but in its procedural transitions: which operation to execute under a given state and which observable outcome to produce.
Based on this observation, we propose State--Move--Outcome (SMO), a compilation-like framework that decomposes raw skills into reusable procedural bricks and organizes admitted bricks as an externalized symbolic structure.
When compiling a skill, SMO first retrieves and composes existing bricks, dynamically connects them through Outcome-to-State conditions, and creates a candidate brick only for an uncovered procedure; the candidate becomes reusable only after validation and admission.
Experiments on SkillsBench v1.1 demonstrate that SMO represents 232 rawSkills from 87 underlying tasks with 95 shared bricks, reducing the corresponding skill representations from 366,472 to 149,154 tokens---a \textbf{59.3\% reduction and $\mathbf{2.46}\times$ compression}.
Across 232 skill-conditioned evaluation instances, SMO raises the average task score from \textbf{0.6379 to 0.7478} and increases 5/5 Pass instances from \textbf{138 to 164}, while matching or outperforming rawSkill on \textbf{208 of 232 instances}.
It also reduces average runtime token consumption by \textbf{38.38\%}.
These results show that compiling natural-language skills into shared procedural units can jointly improve skill reuse, representation efficiency, and downstream execution.
\end{abstract}

\section{Introduction}

Large language model (LLM) agents increasingly rely on procedural knowledge stored outside model parameters.
Prior work has shown that external language structures can be treated not merely as static prompts, but as programs that can be compiled and optimized against downstream objectives \cite{khattab2023dspy}.
Recent agent systems further induce reusable workflows from interaction histories, distill reasoning strategies from successful and failed trajectories, dynamically organize linked memories, and manage growing skill ecosystems \cite{wang2024,ouyang2025,xu2026,li2026a}.
Benchmarks have accordingly begun to evaluate skills as first-class agent artifacts rather than incidental prompts \cite{li2026b}.
Agent skills are therefore becoming persistent procedural infrastructure that must be represented and maintained at corpus scale.

\begin{figure}[t]
\centering
\includegraphics[width=\columnwidth]{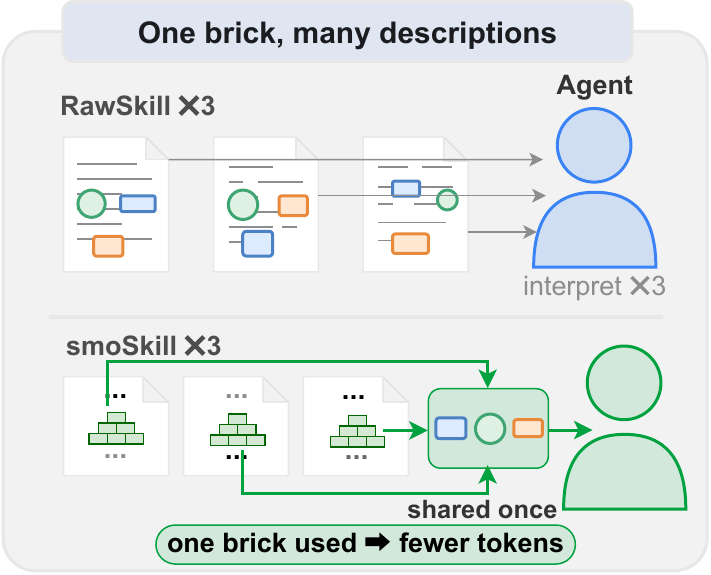}
\caption{One procedure, many descriptions.
Multiple rawSkills may encode the same operation using different natural-language expressions, causing repeated interpretation.
SMO compiles them into smoSkills that reference one shared admitted brick, reducing duplicated procedural context while retaining task-specific composition.}
\label{fig:one}
\end{figure}

Most of this infrastructure is still authored and distributed as natural-language instructions or semi-structured \texttt{SKILL.md} packages.
SkillsBench, for example, defines a skill package around a natural-language \texttt{SKILL.md} file and associated resources; SkillOpt treats such a document as trainable external state; and AgentSkillOS studies ecosystems containing up to hundreds of thousands of complete skills \cite{li2026b,yang2026,li2026a}.
This document-centric representation is convenient for authors, but it obscures reusable procedures.
As illustrated in Figure~\ref{fig:one}, several rawSkills may express the same operation in substantially different prose, requiring the agent to interpret redundant content repeatedly.
A single document may also mix reusable operations with task-specific entities, paths, formats, and explanations.
The resulting corpus is difficult to compare, deduplicate, compose, and update systematically.

\textbf{Existing approaches address adjacent aspects of this problem rather than the corpus-level problem as a whole.} \textbf{(i)}~Experience-learning methods distill workflows, rules, memories, or hierarchical skills from agent trajectories \cite{zhao2024,peng2024,wang2024,chen2024automanual,ouyang2025,xu2026,xia2026,ding2026skillgen}.
\textbf{(ii)}~Compilation and automated-design methods optimize declarative LM programs, function-call plans, agent modules, workflows, or individual skills \cite{khattab2023dspy,kim2023llmcompiler,hu2025automated,zhang2025,shang2025,yang2026,liang2026,chen2026skvm}.
\textbf{(iii)}~Ecosystem-level systems retrieve and orchestrate complete skills, while graph-based tool selectors structure transitions among predefined tools \cite{li2026a,jia2026autotool}.
These directions improve acquisition, optimization, discovery, or runtime execution, but they do not establish a shared procedural basis that can reconstruct many independently authored skill documents.

\textbf{Challenge 1: discovering reusable procedures across heterogeneous descriptions.} Surface similarity is insufficient because distinct wording may implement the same transition, while similar wording may hide different preconditions, operations, or outcomes.
A corpus-level representation must therefore identify equivalence at the level of executable State--Move--Outcome behavior, while separating task-specific values from genuinely new operations.

\textbf{Challenge 2: reconstructing skills through valid composition.} Decomposition alone does not guarantee that the resulting units can be reassembled.
A reusable unit must expose when it applies, what it does, and what it establishes; adjacent units can be connected only when the preceding Outcome satisfies the successor State.
Small residual differences must remain bounded, and uncovered procedures must not be silently approximated by incompatible units.

\textbf{Challenge 3: reducing redundancy without sacrificing utility.} Shorter text is not necessarily a better skill representation.
Corpus compression must preserve the procedures required for downstream tasks, prevent unvalidated units from contaminating the shared basis, and reduce the context repeatedly loaded during execution.
Representation size, task utility, and runtime cost must consequently be evaluated together.

We address these challenges with State--Move--Outcome (SMO), a compilation-like framework for natural-language skill corpora.
SMO atomizes each rawSkill into procedural requirements, reconstructs them through L1 brick matching, L2 instance binding, and L3 bounded-delta patching, and assembles the matched bricks into a graph through Outcome-to-State gates.
Compilation gaps are classified as bindings, bounded deltas, new-brick candidates, or out-of-family cases; admitted candidates and retained deltas are incorporated into the next frozen symbolic-basis snapshot.
On 232 SkillsBench v1.1 skill--task instances, SMO achieves \textbf{$\mathbf{2.46}\times$ representation compression}, raises the average task score from \textbf{0.6379 to 0.7478}, and reduces average runtime token consumption by \textbf{38.38\%}.

Our contributions are:
\begin{enumerate}
\item We introduce a State--Move--Outcome representation that separates reusable procedural transitions from task-specific bindings and bounded differences.
\item We present a corpus-level compilation procedure that prioritizes brick reuse, constructs skill graphs through Outcome-to-State gates, and admits uncovered procedures only through controlled basis evolution.
\item We demonstrate that shared procedural compilation can improve downstream task utility while reducing both skill-representation size and runtime token consumption.
\end{enumerate}

\section{Related Work}

\subsection{Agent Skill Learning and Management}

Recent agents convert interaction histories into reusable external knowledge.
ExpeL allows agents to extract transferable insights from experience, while Hypothesis, Verification, and Induction develops skills through self-driven interaction and verification \cite{zhao2024,peng2024}.
Agent Workflow Memory induces recurring workflows, ReUseIt synthesizes workflows from successful and failed executions, and ReasoningBank distills generalizable reasoning strategies \cite{wang2024,liu2026,ouyang2025}.
AutoManual maintains structured rules before formulating an instruction manual, A-MEM dynamically links and evolves memory notes, and SkillGen induces an action-centric graph of step-wise skills \cite{chen2024automanual,xu2026,ding2026skillgen}.
SkillRL further constructs a hierarchical skill library that co-evolves with the policy \cite{xia2026}.
These methods primarily learn reusable artifacts from interaction trajectories.
SMO instead starts from an existing corpus of independently authored skills and identifies procedural transitions shared across documents.

Other work operates at the skill-document or ecosystem level.
SkillOpt optimizes individual natural-language skills, SkillsBench evaluates modular skill packages, and Anything2Skill turns heterogeneous resources into complete structured skill contracts organized in a SkillBank \cite{yang2026,li2026b,pan2026}.
AgentSkillOS organizes complete skills through a capability tree and DAG pipelines \cite{li2026a}.
These systems establish the value of skill optimization, structuring, evaluation, and orchestration, but retain the complete skill as their principal unit.
SMO instead decomposes many skills into a shared procedural basis and represents each skill through bindings and brick composition.

\subsection{Compilation and Structured Skill Representation}

DSPy establishes a compilation paradigm in which declarative LM programs are optimized against task-level objectives rather than executed as fixed prompts \cite{khattab2023dspy}.
At the agent-system level, ADAS searches over code-defined designs \cite{hu2025automated}.
AgentSquare searches modular combinations \cite{shang2025}, while AFlow searches workflow graphs \cite{zhang2025}.
SMO applies this optimization perspective at a different representation level: it compiles independently authored natural-language skills into reusable State--Move--Outcome transitions shared across skills.

Recent work also compiles or restructures individual agent programs and skills.
LLMCompiler constructs dependency-aware plans for parallel function calling; the Scheduling--Structural--Logical representation converts one skill document into a typed graph; and SkVM compiles skills for portable execution across models and harnesses \cite{kim2023llmcompiler,liang2026,chen2026skvm}.
AutoTool constructs a directed graph over predefined tools to reduce repeated model inference during tool selection \cite{jia2026autotool}.
These systems structure one program, skill package, or tool set at a time.
SMO instead discovers shared procedural units across a skill corpus and composes them only when their Outcome and State conditions are compatible.

\subsection{Compression of External Agent Knowledge}

Compression of external knowledge must preserve downstream utility rather than merely shorten its surface form.
Context-aware Prompt Compression removes less relevant sentences while retaining answer-critical context, CMT compresses documents into memories for continual knowledge learning, and SkillRL abstracts experience into hierarchical skills \cite{liskavets2025,li2025cmt,xia2026}.
SMO instead compiles an entire skill corpus: repeated procedures are stored once as admitted bricks, task-specific differences remain in bindings or bounded deltas, and evaluation jointly measures representation tokens, task performance, and runtime consumption.

\section{Method}

State--Move--Outcome (SMO) compiles an independently authored natural-language skill against a frozen symbolic basis.
For a rawSkill $s_i$ and basis snapshot $\mathcal{B}^{(k)}$, the complete compilation relation is
\begin{equation}
G_i = \mathrm{Compile}\left(s_i; \mathcal{B}^{(k)}\right) = (C_i, E_i, \theta_i, \Delta_i),
\label{eq:compile}
\end{equation}
where smoSkill $G_i$ contains selected bricks $C_i$, topology $E_i$, bindings $\theta_i$, and bounded adaptations $\Delta_i$.
Atomization defines what must be reconstructed; L1--L3 produce $C_i$, $\theta_i$, and $\Delta_i$; and Outcome-to-State composition constructs $E_i$.

\begin{figure*}[t]
\centering
\includegraphics[width=\textwidth]{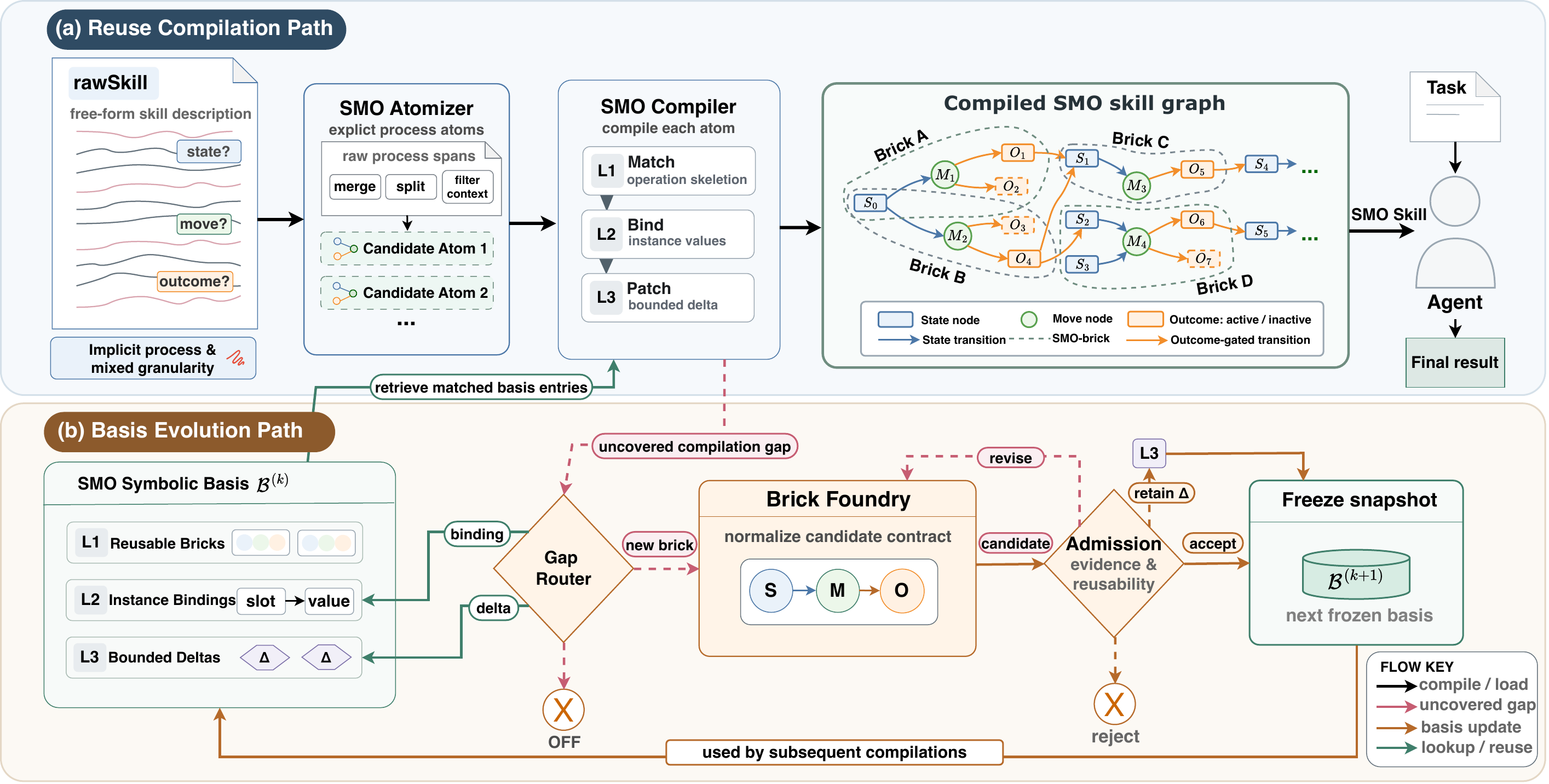}
\caption{Overview of SMO.
(a) The reuse path atomizes a rawSkill, reconstructs its requirements through L1 matching, L2 binding, and L3 bounded patching, and assembles a skill graph for Agent execution.
(b) Only compilation gaps enter basis evolution.
New operations join the next frozen snapshot only after candidate construction and admission; Agent execution does not feed back into SMO.}
\label{fig:two}
\end{figure*}

Figure~\ref{fig:two} separates reuse compilation from basis evolution.
The upper path computes Equation~\ref{eq:compile}; the lower path handles uncovered content.
Accepted content affects only later frozen snapshots, never the current compilation.

\subsection{From rawSkill to Compiled Components}

\textbf{Frozen basis and brick contract.} The snapshot used at round $k$ is $\mathcal{B}^{(k)}=(\mathcal{B}_1^{(k)},\mathcal{B}_2^{(k)},\mathcal{B}_3^{(k)})$.
The layers contain reusable operation skeletons, binding information, and bounded differences.
An L1 brick $b=(S_b,M_b,O_b,P_b,F_b,\nu_b)$ specifies its State, Move, Outcome, bindable slots, failure boundary, and verifier.
State, Move, Outcome, and the failure boundary decide reuse; $P_b$ constrains binding, and $\nu_b$ checks the instantiated result.

\textbf{Atomization.} The Atomizer converts $s_i$ into an ordered set of procedural requirements $R_i=(r_{i,1},\ldots,r_{i,K_i})$.
It filters non-executable context, merges spans describing one transition, splits independent transitions, and makes supported prerequisites and outcomes explicit.
Dependency cues are retained for topology construction.
Requirements specify what Equation~\ref{eq:compile} must cover; they are not automatically admitted as bricks.

\textbf{L1 matching.} For each requirement $r$, semantic retrieval proposes a small L1 candidate set.
A candidate brick $b$ is reusable under a tentative binding $\theta$ only if
\begin{equation}
\begin{aligned}
\mathrm{Match}(r,b,\theta) &= Q_S(r,b[\theta]) \wedge Q_M(r,b[\theta]) \\
&\quad \wedge Q_O(r,b[\theta]) \wedge Q_F(r,b[\theta]).
\end{aligned}
\label{eq:match}
\end{equation}
The predicates check State, operation roles, Outcome coverage, and failure boundaries.
Similarity narrows retrieval but cannot override a failed check.
A matched brick or bounded matched path enters $C_i$.

\textbf{L2 binding and L3 adaptation.} L2 maps the slots of selected bricks to values in the current rawSkill.
For a typed slot $p$, let $\mathrm{Dom}(p)$ be its valid domain, and let $p_a \equiv p_b$ denote two occurrences that must share a value.
The skill-level binding $\theta_i$ must satisfy
\begin{equation}
\begin{gathered}
\theta_i(p) \in \mathrm{Dom}(p), \\
p_a \equiv p_b \Longrightarrow \theta_i(p_a) = \theta_i(p_b).
\end{gathered}
\label{eq:binding}
\end{equation}
This enforces types, units, roles, and repeated-variable consistency without specializing L1.
For example, one sorting brick can bind its field and direction to \emph{price} and \emph{ascending}.

When binding is insufficient, L3 may attach a bounded adaptation $\delta_{i,u}$ to $b_u$, such as a local tie-breaking rule.
The effective brick is $\tilde{b}_{i,u}=b_u[\theta_i]\oplus\delta_{i,u}$, where empty $\delta_{i,u}$ means no adaptation; nonempty adaptations form $\Delta_i$.
This preserves a shared skeleton while allowing small skill-specific differences.
New branches, aggregations, recovery procedures, and persistent state changes instead become compilation gaps.
Every requirement in $R_i$ must be covered by $C_i$, instantiated through $\theta_i$, and, when needed, adapted through $\Delta_i$ before topology construction.

\subsection{Topology Construction and Structural Guards}

SMO connects effective bricks by operational dependencies rather than by copied sentence order.
For two effective bricks $\tilde{b}_{i,u}$ and $\tilde{b}_{i,v}$, define
\begin{equation}
\begin{aligned}
g_{uv} &= \mathrm{Sat}\left(O_{\tilde{b}_{i,u}}, S_{\tilde{b}_{i,v}}\right) \\
&\quad \wedge \neg \mathrm{Conflict}\left(O_{\tilde{b}_{i,u}}, F_{\tilde{b}_{i,v}}\right), \\
(u,v) \in E_i &\Longleftrightarrow g_{uv} = 1.
\end{aligned}
\label{eq:gate}
\end{equation}
An edge therefore requires a predecessor Outcome that satisfies the successor State without violating its failure boundary.
Because the gate follows $\theta_i$ and $\Delta_i$, topology reflects the concrete rawSkill.
It supports sequences, branches, merges, and parallel subprocedures while exposing missing dependencies.

Before emitting $G_i$, the compiler checks that all requirements are covered, selected bricks form a connected executable graph, every branch terminates, and terminal Outcomes cover the requested goal:
\begin{equation}
\begin{aligned}
\mathrm{Valid}(G_i) &= \mathrm{Cover}(R_i,G_i) \wedge \mathrm{Connect}(G_i) \\
&\quad \wedge \mathrm{Terminate}(G_i) \wedge \mathrm{Goal}(s_i,G_i).
\end{aligned}
\label{eq:valid}
\end{equation}
These checks catch omissions, disconnected nodes, unterminated branches, and incomplete results.
They establish structural consistency, not semantic equivalence or guaranteed success.

\begin{figure*}[t]
\centering
\includegraphics[width=\textwidth]{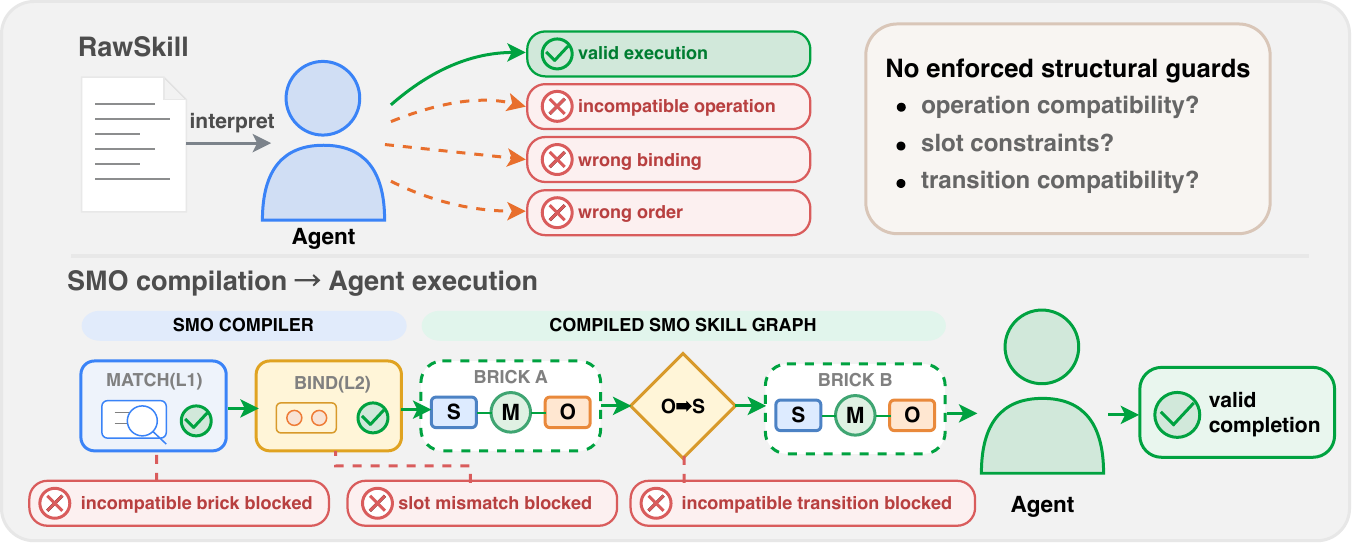}
\caption{Structural guards before Agent execution.
L1 matching blocks incompatible operations, L2 binding blocks slot mismatches, and Outcome-to-State gates block incompatible transitions.
The guards prune identifiable structural errors but do not guarantee task success.}
\label{fig:three}
\end{figure*}

As Figure~\ref{fig:three} illustrates, L1 blocks incompatible operations, L2 blocks invalid bindings, and Outcome-to-State gates block invalid orderings.
L3 adaptations cannot bypass these checks.

\subsection{Controlled Basis Evolution}

The snapshot $\mathcal{B}^{(k)}$ remains immutable.
A requirement not assignable to $C_i$, $\theta_i$, or $\Delta_i$ becomes a gap, classified as binding, bounded delta, possible new brick, or out-of-family.
Binding and delta routes query frozen layers without writing to them; out-of-family procedures are not absorbed.

For a possible new operation, the Brick Foundry constructs candidate $c$ with a complete contract and graph position.
Admission requires completeness, novelty, connectivity, verifiability, and plausible reuse:
\begin{equation}
\begin{aligned}
\mathrm{Admit}(c) &= A_{\mathrm{complete}}(c) \wedge A_{\mathrm{novel}}(c) \wedge A_{\mathrm{connect}}(c) \\
&\quad \wedge A_{\mathrm{verify}}(c) \wedge A_{\mathrm{reuse}}(c).
\end{aligned}
\label{eq:admit}
\end{equation}
Incomplete candidates are revised, local differences may remain deltas, and unsupported candidates are rejected.
Accepted bricks and retained deltas enter the next frozen snapshot $\mathcal{B}^{(k+1)}$ without altering $\mathcal{B}^{(k)}$ or current $G_i$.
Thus only compilation gaps evolve the basis; Agent execution does not feed back.

\subsection{Execution, Reuse, and Complexity}

At runtime, the Agent traverses $E_i$, checks each bound State, performs the Move, uses $\nu_b$ to confirm the Outcome, and follows an enabled successor.
It executes skill-specific $\tilde{b}_{i,u}$ while contracts in $C_i$ remain shared.

For a loaded set $Q$ of smoSkills, repeated bricks are stored once through $U_Q=\bigcup_{i\in Q}C_i$.
With $L(\cdot)$ denoting tokenized representation length, raw and compiled loading costs are
\begin{equation}
\begin{aligned}
C_{\mathrm{raw}}(Q) &= \sum_{i\in Q} L(s_i), \\
C_{\mathrm{SMO}}(Q) &= L(U_Q) + \sum_{i\in Q} L((E_i,\theta_i,\Delta_i)).
\end{aligned}
\label{eq:cost}
\end{equation}
The first term stores repeated bricks once; the second preserves skill-specific topology, bindings, and adaptations.
Savings combine compact reconstruction with cross-skill deduplication and vanish at the sharing level when no bricks recur.

For complexity, let $T_i$ be the length of $s_i$, $K_i$ its number of requirements, $h$ the maximum number of retrieved candidates per requirement, and $\ell$ the maximum tested short-path length.
After indexed retrieval, the main costs satisfy
\begin{equation}
\begin{aligned}
T_{\mathrm{atom}} &= O(T_i), & T_{\mathrm{recon}} &= O(K_i h^\ell), \\
T_{\mathrm{graph}} &= O(|C_i|^2), & T_{\mathrm{run}} &= O(|C_i|+|E_i|).
\end{aligned}
\label{eq:complexity}
\end{equation}
Atomization is linear in rawSkill length, bounded path testing dominates reconstruction, topology construction is quadratic in selected bricks, and traversal is linear in the compiled graph, excluding model calls and tools.
Derivations are deferred to the supplement.

\section{Experiments}

We evaluate whether SMO can compile natural-language skills into a compact shared representation while improving task performance and reducing inference cost.
The main study covers 87 SkillsBench tasks and 232 associated skills; a paired diagnostic corpus separately controls for generic structuring.
We examine representation compression, reuse, downstream performance, inference tokens, and ablations of structure and sharing.

\subsection{Experimental Setup}

\textbf{Dataset and raw skills.} We use SkillsBench v1.1 \cite{li2026b} at commit \texttt{a8eefb4}: 87 tasks with isolated environments and 232 official \texttt{SKILL.md} documents.
We compile every document into one smoSkill while preserving its task mapping, yielding 232 skill-conditioned instances over 87 distinct tasks.

\textbf{Controlled fixture corpus.} To separate generic structuring from brick sharing, we construct an internal engineering-diagnostic corpus with 20 source clusters and 60 primary units.
A source cluster groups units derived from the same diagnostic source; a primary unit is the content-preserving item transformed across conditions.
Each unit is converted one-to-one into three representations without changing its content or order, yielding 60 paired units per condition and 180 test units in total.
The paired design varies the external representation while holding unit content and ordering fixed.
These fixtures are used only for this representation-level control and are not drawn from SkillsBench, TerminalWorld, or Anything2Skill's official data.

\textbf{Baselines and ablations.} On SkillsBench, No Skill uses only the task instruction, rawSkill loads the official document, and SMO loads its compiled counterpart.
Length-Matched Compression controls for text length, while SMO w/o Sharing assigns private bricks to control for cross-skill reuse.
All use identical task inputs, executor, and environment.

On the fixture corpus, Raw Control preserves the unstructured source; Structured Contract uses GPT-5.5 to render the same content as a complete standalone contract inspired by Anything2Skill \cite{pan2026}; and Full SMO uses its already compiled brick decomposition.
A local deterministic renderer serializes the SMO objects; it neither discovers nor rewrites bricks and makes no LLM call.
Structured Contract controls for generic structuring; it is not a reproduction of Anything2Skill's taxonomy, SkillBank, retrieval pipeline, or datasets.

\textbf{Implementation details.} Task execution uses the configured model identifier \texttt{openai/gpt-5.5} with the Codex-ACP agent and \texttt{reasoning\_effort=high}.
We use BenchFlow 0.4.0 (commit \texttt{7a479e9}) in isolated Docker environments.
Each skill-conditioned instance is executed independently five times, with identical decoding settings across the compared methods.
We retain the task-specific time limits declared by the benchmark configurations rather than imposing a single global timeout.

Representation length is computed from the serialized representation with the same tokenizer implementation pinned in the evaluation code for every condition.
It includes only the rawSkill or smoSkill representation, excluding task inputs and execution traces.
Runtime token consumption is obtained from execution logs.
We first average the five runs of each instance and then report a macro-average over the relevant instances.
This metric measures task-execution cost after deployment and excludes the one-time offline cost of compiling skills with SMO.

\textbf{Evaluation metrics.} For each run, an independent GPT-5.5 judge scores the complete execution trace in $[0,1]$ under a fixed rubric.
The judge provides semantic scores; deterministic programs check format and constraints and aggregate results.
On the fixture corpus, a unit is accepted only when its semantic score meets the pre-specified criterion and every deterministic check succeeds; the criterion is fixed across conditions.
We average five run-level scores per skill-conditioned instance and then macro-average over 232 instances.
\emph{5/5 Pass} requires all five runs to be fully successful and is therefore not pass@5.
Better/Tie/Worse compares instance-level averages.

Representation efficiency is reported through the number of shared bricks, total skill-representation tokens, percentage token reduction, and compression ratio.
Token reduction is the percentage decrease from the rawSkill representation to the corresponding smoSkill representation, and the compression ratio divides the rawSkill token count by the smoSkill token count.
Runtime efficiency is summarized by the average execution tokens per instance and their percentage decrease relative to rawSkill.
These quantities are described directly because they are standard ratios rather than new metrics introduced by SMO.

\begin{table}[t]
\small
\setlength{\tabcolsep}{1.5pt}
\centering
\caption{Representation results.
Panel (b) compares the two structured conditions over 60 paired units each and normalizes indices to Structured Contract.}
\label{tab:representation}
\begin{tabular}{lccccc}
\multicolumn{6}{c}{\textit{(a) SkillsBench corpus compression}} \\
\toprule
Representation & Skills & Shared bricks & Tokens & Reduction & Ratio \\
\midrule
rawSkill & 232 & -- & 366,472 & 0.0\% & $1.00\times$ \\
\textbf{smoSkill} & \textbf{232} & \textbf{95} & \textbf{149,154} & \textbf{59.3\%} & \textbf{$2.46\times$} \\
\bottomrule
\end{tabular}

\vspace{0.6em}

\begin{tabular}{lcccc}
\multicolumn{5}{c}{\textit{(b) Controlled structured-representation comparison}} \\
\toprule
Representation & Tok. idx. $\downarrow$ & Comp. $\uparrow$ & Byte idx. $\downarrow$ & Reuse $\uparrow$ \\
\midrule
Structured Contract & 1.0000 & $1.00\times$ & 1.0000 & -- \\
\textbf{Full SMO} & \textbf{.5337} & \textbf{$1.87\times$} & \textbf{.4646} & \textbf{95.0\%} \\
\bottomrule
\end{tabular}
\end{table}

\section{Skill Representation Compression and Reuse}

Table~\ref{tab:representation} reports corpus-level representation efficiency.
The 232 rawSkills contain 366,472 representation tokens in total.
SMO compiles them into 232 corresponding smoSkills using 95 shared bricks, reducing the representation to 149,154 tokens.
This amounts to a 59.3\% reduction and a $2.46\times$ compression ratio.
The comparison measures the token length of functionally corresponding skill representations rather than directly comparing the number of rawSkills with the number of bricks.

\textbf{Controlled structured baseline.} The fixture verifier accepts 36/60 Raw Control units and 60/60 units under both structured conditions.
Holding acceptance fixed, Table~\ref{tab:representation}(b) shows that Full SMO reduces the context-token index by 46.63\% and serialized bytes by 53.54\% relative to Structured Contract, giving $1.87\times$ token compression.
Unique-brick reuse is the fraction of evaluated bricks referenced by at least two distinct source clusters; the 95.0\% rate therefore measures cross-cluster sharing.

The 36/60 versus 60/60 acceptance contrast is not used as a downstream task-performance claim; it verifies that the two structured conditions preserve the fixture contracts before their representation sizes are compared.
Raw Control is therefore omitted from Table~\ref{tab:representation}(b): comparing its size with accepted representations would confound compression with missing contract content.
Token and byte indices are computed from the exact serialized outputs of each condition.
The token index measures model-context load, whereas the byte index checks that the observed reduction is not specific to one tokenizer.

The reuse metric is inventory-based rather than invocation-weighted: each distinct brick counts once, so one frequently used brick cannot dominate the result.
Requiring references from at least two source clusters also excludes repeated use confined to one source.
The metric therefore targets the cross-source structural sharing that distinguishes a brick inventory from a collection of independent contracts.

Because the 60 primary units are transformed one-to-one, every Full SMO item has a content-matched Structured Contract counterpart.
The 46.63\% token reduction is therefore measured over paired representations rather than unrelated corpora or different task mixtures.
We treat this result as a controlled representation comparison after both conditions meet the same acceptance rule, not as an end-to-end reproduction or task-accuracy comparison with Anything2Skill.

Together with the SkillsBench ablations, the controls isolate three alternatives.
Length-Matched Compression tests whether fewer tokens alone suffice; Structured Contract tests whether a complete standalone schema suffices; and SMO without sharing tests whether private structured units suffice.
Full SMO is the only condition that combines procedural decomposition with cross-skill sharing.

\section{Task Performance on SkillsBench}

Table~\ref{tab:performance} compares No Skill, rawSkill, and SMO.
Loading the official rawSkills raises the average task score from 0.4397 to 0.6379 and increases the number of 5/5 Pass instances from 95 to 138, showing that the official skills provide useful task knowledge.
SMO further raises the average score to 0.7478 and achieves 5/5 Pass on 164 instances.
Relative to rawSkill, this corresponds to an absolute average-score improvement of 0.1099 and a net increase of 26 5/5 Pass instances.

At the instance level, SMO is better than, tied with, or worse than rawSkill on 50, 158, and 24 instances, respectively.
Fifty instances move from non-5/5 to 5/5 Pass, whereas 24 move in the opposite direction.
Overall, SMO matches or outperforms rawSkill on 208 of 232 instances.
The regressions nevertheless show that structured compilation is not lossless for every skill.

\begin{table}[t]
\small
\setlength{\tabcolsep}{1.5pt}
\centering
\caption{Task performance on SkillsBench.
B/T/W is relative to rawSkill.}
\label{tab:performance}
\begin{tabular}{lccc}
\toprule
Method & Avg. score $\uparrow$ & 5/5 Pass $\uparrow$ & B/T/W \\
\midrule
No Skill & 0.4397 & 95/232 & -- \\
rawSkill & 0.6379 & 138/232 & Reference \\
\textbf{SMO} & \textbf{0.7478} & \textbf{164/232} & \textbf{50/158/24} \\
\bottomrule
\end{tabular}
\end{table}

\begin{table}[t]
\small
\setlength{\tabcolsep}{1.5pt}
\centering
\caption{Ablations.
Tokens are in thousands (k); 5/5 Pass is a count out of 232.}
\label{tab:ablations}
\begin{tabular}{lccc}
\multicolumn{4}{c}{\textit{(a) Task quality}} \\
\toprule
Method & Avg. score $\uparrow$ & 5/5 $\uparrow$ & B/T/W \\
\midrule
rawSkill & 0.6379 & 138 & Reference \\
Length-matched & 0.6025 & 129 & 20/183/29 \\
SMO w/o sharing & 0.7494 & 164 & 49/160/23 \\
\textbf{Full SMO} & \textbf{0.7478} & \textbf{164} & \textbf{50/158/24} \\
\bottomrule
\end{tabular}

\vspace{0.6em}

\begin{tabular}{lccccc}
\multicolumn{6}{c}{\textit{(b) Representation and runtime efficiency}} \\
\toprule
Method & Bricks $\downarrow$ & Repr. k $\downarrow$ & Save $\uparrow$ & Run k $\downarrow$ & Save $\uparrow$ \\
\midrule
rawSkill & -- & 366.5 & 0.0\% & 1,296.8 & 0.0\% \\
Length-matched & -- & 149.2 & 59.3\% & 1,207.0 & 6.93\% \\
SMO w/o sharing & 379 & 228.3 & 37.7\% & 1,190.9 & 8.17\% \\
\textbf{Full SMO} & \textbf{95} & \textbf{149.2} & \textbf{59.3\%} & \textbf{799.1} & \textbf{38.38\%} \\
\bottomrule
\end{tabular}
\end{table}

\begin{table}[t]
\small
\setlength{\tabcolsep}{1.5pt}
\centering
\caption{Average inference tokens (thousands) by 5/5 Pass outcome.}
\label{tab:tokens}
\begin{tabular}{lcccc}
\toprule
Outcome group & $n$ & rawSkill & SMO & Save $\uparrow$ \\
\midrule
Both 5/5 & 114 & 1,105.3 & 744.7 & 32.63\% \\
Neither 5/5 & 44 & 1,244.9 & 784.2 & 37.01\% \\
SMO only & 50 & 1,420.9 & 740.6 & 47.88\% \\
rawSkill only & 24 & 2,043.2 & 1,207.1 & 40.92\% \\
All & 232 & 1,296.8 & 799.1 & 38.38\% \\
\bottomrule
\end{tabular}
\end{table}

\subsection{Inference Efficiency}

To separate token use from the composition of successful instances, Table~\ref{tab:tokens} partitions the 232 pairs by whether rawSkill and SMO attain 5/5 Pass.
SMO reduces average inference tokens in every group: by 32.63\% when both methods pass and by 47.88\% for SMO-only successes.
Thus, its performance improvement is not obtained by spending more inference tokens.

Across all pairs, SMO reduces average inference cost from 1,296,830.6 to 799,111.3 tokens per instance, saving 38.38\%.
It also saves 40.92\% on the 24 regressions, showing that efficiency gains and task regressions can coexist; token length alone does not explain a regression, so task performance and runtime cost must be reported jointly.

\subsection{Ablation Study}

Table~\ref{tab:ablations} isolates representation length, SMO structure, and cross-skill reuse.
Despite matching full SMO's 149,154 tokens, Length-Matched Compression scores 0.6025 with 129 5/5 Pass instances, below rawSkill.
Text shortening alone therefore cannot explain SMO's gains.

The fixture comparison adds a stronger structural control: after both Structured Contract and Full SMO reach 60/60 verifier acceptance, shared bricks still reduce context tokens by 46.63\%.
This separates SMO's sharing benefit from generic contract structuring.

Disabling cross-skill sharing increases the representation to 379 private bricks and 228,312 tokens, reducing representation savings from 59.3\% to 37.7\% and runtime savings from 38.38\% to 8.17\%.
Yet SMO without sharing and full SMO score 0.7494 and 0.7478, respectively, and both attain 5/5 Pass on 164 instances.
Brick sharing therefore primarily improves representation and runtime efficiency without an evident aggregate-quality loss.

When a task requires multiple smoSkills, a shared brick is injected once; private copies re-enter the context through each skill.
Taken together, the controls show that shorter text alone cannot account for the gains: SMO structure preserves task utility, while cross-skill sharing reduces representation and runtime context.

The near-equal aggregate scores of full SMO and the no-sharing variant support this division of roles, but do not imply strict causal independence: changing context length or ordering may still alter individual executions.
These ablations evaluate deployed representations and runtime traces; the separate one-time compilation cost is not included.

\subsection{Limitations}

Our evaluation has two main limitations.
First, it uses a fixed SkillsBench collection, one executor configuration (GPT-5.5 with Codex-ACP), and a GPT-5.5 judge under a fixed rubric.
The Structured Contract comparison uses an internal diagnostic fixture corpus rather than Anything2Skill's official data or complete system.
Results may not transfer directly to other models, harnesses, or skill collections.

Second, SMO does not improve every instance: 24 instances regress from 5/5 Pass to non-5/5, suggesting that compilation may occasionally omit useful details or alter how instructions are interpreted.
Moreover, the reported runtime tokens exclude the one-time offline compilation cost, so the overall benefit depends on how frequently compiled skills are reused.

\section{Conclusion}

We introduced State--Move--Outcome (SMO), a compilation-like framework that transforms natural-language skills into reusable procedural bricks and dynamically assembles them into executable skill graphs.
On SkillsBench v1.1, SMO reduced representation tokens by 59.3\%, improved the average task score from 0.6379 to 0.7478, and reduced average runtime token consumption by 38.38\%.
Ablations indicate that these gains arise from the SMO structure and cross-skill reuse rather than text shortening or standalone structuring alone.

Future work will test additional models and skill collections while strengthening admission and measuring the amortized benefit of reuse.

\bibliography{aaai2027}

\end{document}